\documentclass[sigconf]{acmart}
\AtBeginDocument{%
  }

\setcopyright{none}

\AtBeginDocument{%
  \fancyhead[RO]{\sffamily\footnotesize\shortauthors}%
  \fancyhead[LE]{}%
}

\usepackage{multirow}
\usepackage[most]{tcolorbox}
\usepackage{booktabs}
\usepackage{graphicx}

\begin{document}

%%
%% The "title" command has an optional parameter,
%% allowing the author to define a "short title" to be used in page headers.
\title{Multimodal Adaptive Expert Selection with Text Routing and Ordinal Prototype Optimization for Sentiment Analysis}

%%
%% The "author" command and its associated commands are used to define
%% the authors and their affiliations.
%% Of note is the shared affiliation of the first two authors, and the
%% "authornote" and "authornotemark" commands
%% used to denote shared contribution to the research.
\author{Xiaode Chen}
\authornote{Both authors contributed equally to this research.}
\affiliation{%
  \institution{Jianghan University}
  \city{Wuhan}
  \country{China}
}
\email{chenxiaode2003@stu.jhun.edu.cn}

\author{Jiakang Yu}
\authornotemark[1]
\affiliation{%
  \institution{Jianghan University}
  \city{Wuhan}
  \country{China}
}
\email{jiakangyu@stu.jhun.edu.cn}

\author{Hongtao Deng}
\authornote{Corresponding authors.}
\affiliation{%
  \institution{Jianghan University}
  \city{Wuhan}
  \country{China}
}
\email{hongtaodeng@jhun.edu.cn}

\author{Huina Qu}
\affiliation{%
  \institution{Jianghan University}
  \city{Wuhan}
  \country{China}
}
\email{quhuina2004@gmail.com}

\author{Xun Zhu}
\affiliation{%
  \institution{Jianghan University}
  \city{Wuhan}
  \country{China}
}
\email{zhuxun@jhun.edu.cn}

\author{Yinxia Lou}
\affiliation{%
  \institution{Jianghan University}
  \city{Wuhan}
  \country{China}
}
\email{yinxia@jhun.edu.cn}

%%
%% By default, the full list of authors will be used in the page
%% headers. Often, this list is too long, and will overlap
%% other information printed in the page headers. This command allows
%% the author to define a more concise list
%% of authors' names for this purpose.
\renewcommand{\shortauthors}{Xiaode et al.}

%%
%% The abstract is a short summary of the work to be presented in the
%% article.
\begin{abstract}
Multimodal Sentiment Analysis (MSA) is a fundamental component of affective computing that aims to decipher complex emotional states by integrating verbal content with non-verbal cues including vocal intonation and facial micro-expressions. While recent disentanglement-based approaches have advanced the field, their potential is hindered by two methodological challenges. First, static computation graphs process all samples indiscriminately regardless of semantic complexity, which leads to suboptimal representation for diverse emotional expressions and contextual scenarios. Second, generic contrastive objectives often neglect the intrinsic ordinal hierarchy of sentiment intensities. To systematically address these limitations, we introduce \textbf{M}ultimodal \textbf{A}daptive \textbf{E}xpert \textbf{S}election with \textbf{T}ext \textbf{R}outing and \textbf{O}rdinal prototype optimization (\textbf{MAESTRO}), a novel framework designed to dynamically orchestrate and refine multimodal representations. Drawing inspiration from an orchestra conductor, we design a Text-Guided Hybrid Mixture-of-Experts (MoE) mechanism. Unlike static fusion, this module utilizes linguistic context as a routing signal to dynamically activate specific audio-visual experts, thereby resolving cross-modal ambiguity through adaptive feature enhancement. Furthermore, to capture fine-grained sentiment gradations, we propose an Ordinal-aware Prototype Contrastive Learning (O-PCL). By incorporating distance-based penalties into the prototype learning objective, O-PCL enforces a structured latent space that preserves the natural order of emotion. Extensive experiments on the CMU-MOSI and CMU-MOSEI benchmarks demonstrate that MAESTRO achieves state-of-the-art performance, and qualitative analysis further confirms the interpretability of our dynamic routing paradigm.
\end{abstract}

%%
%% The code below is generated by the tool at http://dl.acm.org/ccs.cfm.
%% Please copy and paste the code instead of the example below.
%%


%%
%% Keywords. The author(s) should pick words that accurately describe
%% the work being presented. Separate the keywords with commas.
%% A "teaser" image appears between the author and affiliation
%% information and the body of the document, and typically spans the
%% page.

% \received{20 February 2007}
% \received[revised]{12 March 2009}
% \received[accepted]{5 June 2009}

%%
%% This command processes the author and affiliation and title
%% information and builds the first part of the formatted document.
\maketitle

\section{Introduction}

Multimodal Sentiment Analysis (MSA) stands at the forefront of modern artificial intelligence research with the primary objective of gauging emotional intensity through the synergistic integration of heterogeneous data streams encompassing language, vision, and acoustics. In contrast to conventional unimodal approaches that depend exclusively on linguistic semantics, MSA seeks to capture the multifaceted nature of human communication. Non-verbal signals including voice modulation and subtle facial movements frequently serve as the key to unlocking the genuine intent of a speaker \cite{morency2011towards, baltruvsaitis2018multimodal, zhu2023multimodal, das2023multimodal}. Consequently, this technology offers immense utility in a wide array of applications ranging from personalized recommendation systems in social media to automated monitoring in mental healthcare.

% Figure 1 placement
\begin{figure*}[t]
  \centering
  \includegraphics[width=1.0\textwidth]{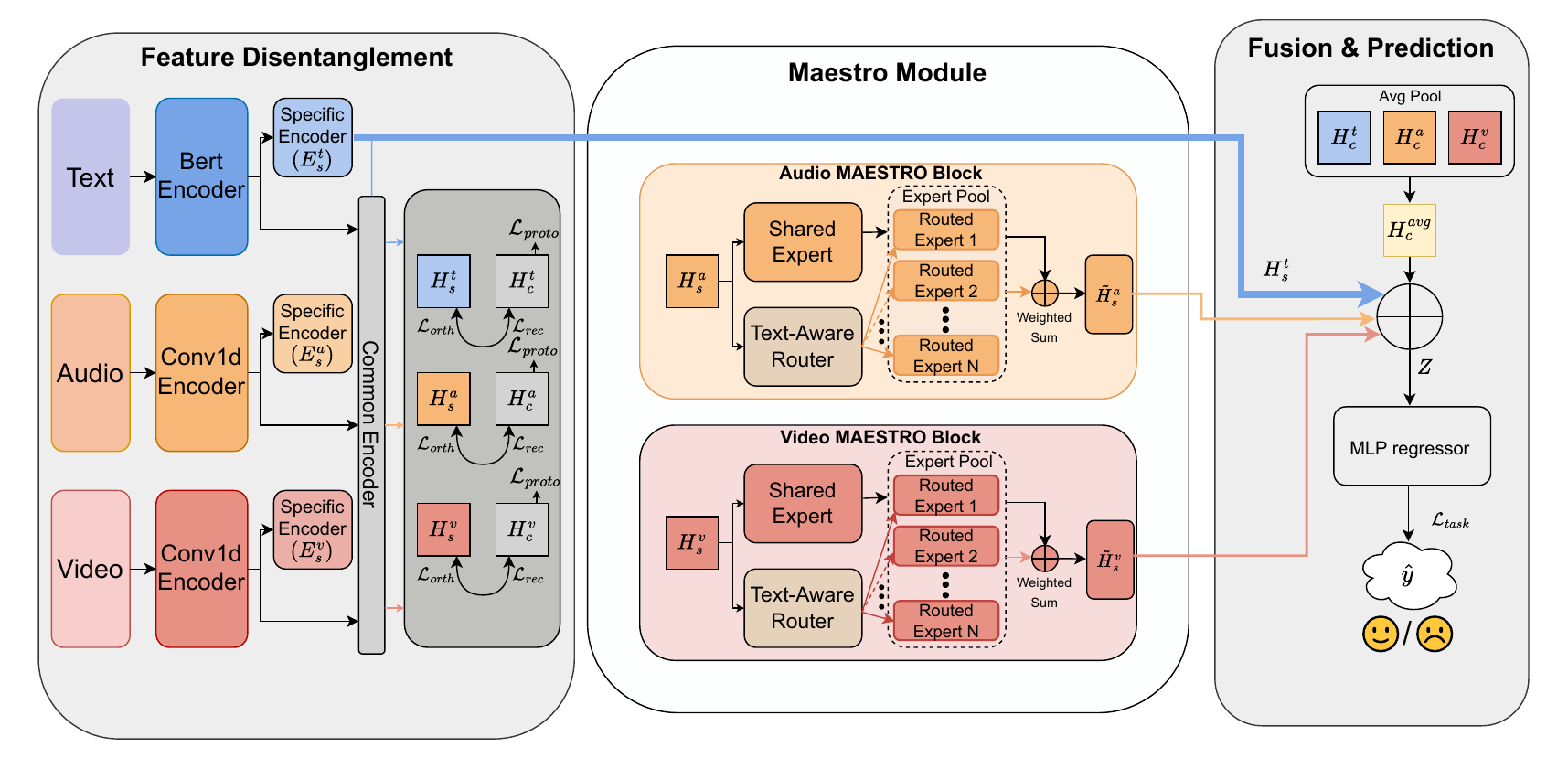} 
  \caption{The overall architecture of MAESTRO, which utilizes a Text-Guided Hybrid MoE mechanism to disentangle modality-specific features, orchestrate expert selection via a "conductor" signal, and fuse enhanced representations for robust sentiment analysis.}
  \label{fig:maestro_arch}
\end{figure*}

To tackle the heterogeneity gap across modalities, the field has witnessed an evolution from early fusion architectures to disentangled representation learning. Seminal fusion works such as TFN \cite{zadeh2017tensor} and MulT \cite{tsai2019multimodal} leveraged tensor fusion or cross-modal Transformers \cite{vaswani2017attention} to capture interactions directly from raw features. Subsequently, to mitigate information redundancy and conflict, the paradigm shifted towards feature disentanglement. State-of-the-art methods like MISA \cite{hazarika2020misa} and DLF \cite{wang2025dlf} decompose multimodal features into modality-invariant and modality-specific subspaces by employing geometric constraints to refine representations. Due to their robustness and compact architecture, these disentanglement-based approaches have established themselves as the dominant paradigm for efficient MSA.

However, despite these advances, existing disentanglement frameworks face two critical limitations in the context of advanced multimedia retrieval.
\begin{enumerate}
    \item \textbf{Inflexible Static Interaction}: Current models rely on static computation graphs that process every sample through identical fixed parameters regardless of its semantic complexity. This "one-size-fits-all" mechanism lacks the adaptability to handle cross-modal variations effectively. For instance, a linguistically explicit sentiment requires minimal verification, whereas an ambiguous sarcastic utterance demands the activation of specialized acoustic or visual modules to disambiguate the intent. Static models lack the flexibility to adapt to such distinct reasoning needs.
    
    \item \textbf{Neglect of Ordinal Semantics}: Existing methods often treat sentiment analysis as a standard classification or regression problem while ignoring the intrinsic ordinal nature of sentiment intensities, which typically range from strongly negative to strongly positive, with varying levels of intensity. They usually employ generic contrastive losses that treat all negative samples equally. Consequently, they fail to penalize the semantic distance between labels, such as the significant error of misclassifying a strongly positive sample as strongly negative compared to a minor deviation. This lack of ordinal awareness leads to a latent space that is loosely clustered and lacks fine-grained discriminative power.
\end{enumerate}

To address these limitations, we draw inspiration from an orchestra conductor. In a symphony, the conductor dynamically cues different instrument sections based on the score. Similarly, we argue that Language, as the dominant semantic modality, should act as the "conductor" to orchestrate non-verbal modalities. Based on this insight, we propose \textbf{M}ultimodal \textbf{A}daptive \textbf{E}xpert \textbf{S}election with \textbf{T}ext \textbf{R}outing and \textbf{O}rdinal prototype optimization (\textbf{MAESTRO}). To the best of our knowledge, MAESTRO is the first framework to incorporate a Text-Guided Hybrid \textbf{Mixture-of-Experts (MoE)} mechanism within the multimodal disentanglement paradigm. Specifically, we replace static fusion layers with a dynamic routing strategy where the textual context gates and weights specific audio-visual experts to enable sample-specific computation. Furthermore, to structure the semantic space, we introduce a novel \textbf{Ordinal-aware Prototype Contrastive Learning (O-PCL)}. Unlike standard contrastive methods, O-PCL incorporates a distance-based penalty that pushes samples further away from emotionally distant prototypes, thereby enforcing a structured embedding space that aligns with the ordinal hierarchy of sentiment labels.

Our main contributions are summarized as follows:
\begin{itemize}
    \item We propose MAESTRO, a novel framework that breaks the static interaction bottleneck of traditional disentanglement methods by introducing a Language-Guided Dynamic Routing strategy.
    \item We design a Text-Guided Hybrid MoE module that acts as a "conductor," allowing textual semantics to dynamically select and modulate audio-visual experts for adaptive feature enhancement.
    \item We devise an O-PCL that dynamically adjusts penalties based on the semantic distance between sentiment intensities, ensuring that the feature space is aligned with the fine-grained ordinal structure of human emotions.
    \item Extensive experiments on the CMU-MOSI and CMU-MOSEI benchmarks demonstrate that MAESTRO achieves state-of-the-art performance across most key metrics, yielding substantial improvements over strong baselines.
\end{itemize}

\section{Related Work}
\label{sec:related_work}

\subsection{Multimodal Sentiment Analysis}
The field of MSA has evolved from early feature-level fusion to sophisticated deep representation learning. 
Early feature-level fusion methods typically extract unimodal features and explicitly combine them into a joint representation via predefined fusion operators. 
Representative tensor-based approaches, such as TFN \cite{zadeh2017tensor} and LMF \cite{liu2018efficient}, model inter-modal interactions through tensor outer products and low-rank factorization, respectively, offering a direct yet often rigid way to capture cross-modal dependencies. 
In contrast, deep representation learning methods aim to learn fused multimodal representations in an end-to-end manner, where cross-modal interactions are implicitly modeled within neural architectures. 
With the advent of the Transformer, attention-based models became mainstream: MulT \cite{tsai2019multimodal} leverages cross-modal attention to align unaligned sequences, while MISA \cite{hazarika2020misa} learns more robust representations by disentangling modality-invariant and modality-specific factors. 
More recently, advanced frameworks further enhance representation learning and fusion through hierarchical mechanisms \cite{tang2025hierarchical, wen2025hgatt}, knowledge-guided learning \cite{feng-etal-2024-knowledge, yu-etal-2023-conki, li2024multimodal}, adaptive dual-branch networks \cite{geng2023dual, su2025adaptive}, and multi-scale convolutional fusion with contrastive learning \cite{yu2026amccl}. Meanwhile, the significance of text-guided cross-modal interaction extends beyond sentiment analysis: implicit hate speech detection calls for resolving the incongruity between literal language and non-verbal cues \cite{liu2026queryguided}, and multimodal named entity recognition benefits from aligning visual semantics with language \cite{yu2026codemner}.

However, most existing methods rely on static computation graphs where all input samples pass through the same fixed encoders and fusion layers. This "one-size-fits-all" paradigm struggles to handle diverse noise levels and semantic ambiguities in real-world multimodal data. For example, resolving sarcasm or context-dependent non-verbal cues often calls for instance-adaptive cross-modal reasoning rather than fixed interactions, leading to suboptimal performance on hard samples under static architectures.

\subsection{Dynamic Computation and MoE}
Dynamic neural networks adapt their structures or parameters based on the input instance, offering a promising solution to the limitations of static models. Among these, the MoE architecture has gained significant traction in natural language processing and computer vision \cite{shazeer2017outrageously, fedus2022switch,dai2024deepseekmoe}. MoE scales model capacity by activating only a sparse subset of experts for each input token which achieves high efficiency.

Despite its success in other fields, the application of MoE in MSA remains underexplored. While some recent works have begun to introduce MoE-style gating for multimodal fusion \cite{chen2025mixture}, their expert selection is typically conditioned on multimodal token representations and is not explicitly guided by linguistic context as a global routing signal. In contrast, our MAESTRO framework incorporates a Text-Guided Router, which leverages language as a “conductor” to orchestrate visual and acoustic experts, thereby resolving cross-modal ambiguity through dynamic, context-aware expert selection.

\subsection{Contrastive Learning and Ordinal Regression}
Contrastive learning has emerged as a powerful technique for learning distinctive representations by pulling positive pairs closer and pushing negative pairs apart. In the context of MSA, Self-MM \cite{yu2021learning} utilized unimodal label generation and contrastive constraints to refine modality representations. HyCon \cite{mai2022hybrid} further explored hybrid contrastive strategies to capture both intra- and inter-modal dependencies \cite{yang2023confede, yang2024clgsi, khosla2020supervised}.

Nevertheless, standard contrastive learning approaches typically treat sentiment analysis as a categorical classification problem. This approach ignores the intrinsic ordinal nature of sentiment intensities such as the semantic difference between +1 and +2. Such limitations often result in regression errors where predictions violate the natural order of sentiment. To address this, ordinal regression techniques \cite{chu2007support,lian2020context} have been proposed, but few integrate them effectively with contrastive learning. Our proposed O-PCL fills this gap by injecting ordinal constraints into the latent space, ensuring that the learned embeddings respect the hierarchy of sentiment intensities.

\section{Methodology}

\subsection{Overall Architecture}
The framework of the proposed MAESTRO is illustrated in Figure~\ref{fig:maestro_arch}. 
It follows a five-stage pipeline: Multimodal Feature Extraction projects the pre-extracted text, audio, and visual inputs into high-level representations; 
Latent Feature Disentanglement decomposes each modality representation into a modality-specific subspace and a modality-invariant common subspace; 
Text-Guided Feature Enhancement employs the Maestro block to route and refine non-verbal representations conditioned on the text-specific feature; 
Multimodal Fusion aggregates the enhanced modality-specific features together with the common representations; 
and Sentiment Prediction outputs the final sentiment intensity via a regression head.

Given an input video clip, we denote the raw modalities as $\mathcal{X} = \{X_t, X_a, X_v\}$. Initially, we take the pre-extracted modality features stored in \texttt{.pkl} files as inputs and project them into high-level representations $\mathcal{F} = \{E_t, E_a, E_v\}$ using modality-specific input encoders. Specifically, $E_t \in \mathbb{R}^{d_t}$ is obtained via a BERT-based text encoder \cite{devlin2019bert}. For the acoustic and visual streams, the input feature sequences are first mapped to a unified hidden space through lightweight 1D convolutional projection layers and then encoded by Transformer-based networks to produce $E_a \in \mathbb{R}^{d_a}$ and $E_v \in \mathbb{R}^{d_v}$.

To address information heterogeneity and redundancy, the feature disentanglement module decomposes each $E_m$ into a modality-specific subspace $H_s^m$ and a modality-invariant common subspace $H_c^m$ for $m \in \{t, a, v\}$. To ensure the purity and complementarity of these latent spaces, three constraints comprising orthogonality ($\mathcal{L}_{orth}$), reconstruction ($\mathcal{L}_{rec}$), and the proposed O-PCL are incorporated as regularization terms.

The defining component of the MAESTRO framework is the Maestro block, which implements a Text-Guided Hybrid MoE mechanism. It treats the text-specific feature $H_s^t$ as an anchor to orchestrate a shared expert and an expert pool for audio and video enhancement. This process produces refined representations $\tilde{H}_s^a$ and $\tilde{H}_s^v$ that are sensitive to the linguistic context. Finally, the enhanced features are aggregated via a Final Fusion Layer to predict the sentiment intensity $\hat{y}$.

\subsection{Feature Disentanglement}
\label{sec:disentanglement}

A fundamental challenge in multimodal sentiment analysis is the inherent heterogeneity of the data. Raw feature representations typically entail an entanglement of modality-invariant semantics representing the underlying sentiment and modality-specific characteristics such as acoustic background or visual style. Directly fusing these entangled features may introduce redundancy and noise which hinders effective cross-modal interaction. To address this, we propose a geometry-guided disentanglement paradigm. As illustrated in the Feature Disentanglement module of Figure \ref{fig:maestro_arch}, our objective is to mathematically decompose the input features into two independent subspaces: a specific space for unique modal dynamics and a common space for shared sentiment semantics.

\subsubsection{Dual-Stream Feature Decomposition}
Let the high-level feature of modality $m \in \{t,a,v\}$ be denoted as $E_m$. To disentangle the representation, we employ two parallel encoders: a modality-specific encoder $E_s^m$ and a modality-invariant encoder $E_c^m$.
\begin{equation}
    H_s^m = E_s^m\!\left(E_m\right), \qquad H_c^m = E_c^m\!\left(E_m\right),
\end{equation}
Here, $H_s^m$ encodes the exclusive attributes of modality $m$, while $H_c^m$ is intended to capture the sentiment consistency shared across modalities.

\subsubsection{Latent Space Regularization}
 To validate this decomposition and prevent trivial solutions, we impose three geometric regularization constraints.
 
\noindent \textbf{(1) Integrity via Reconstruction.}
To avoid losing information during disentanglement, we regularize the decomposition by requiring the reconstructed feature to stay close to the input feature.
Specifically, we introduce a decoder $D_m$ for each modality $m$. The decoder maps the concatenated representation $H_s^m \oplus H_c^m$ back to the feature space, and we minimize the squared $\ell_2$ reconstruction error:
\begin{equation}
\mathcal{L}_{rec} = \sum_{m \in \{t, a, v\}} \left\| E_m - D_m\!\left( H_s^m \oplus H_c^m \right) \right\|_2^2,
\end{equation}
Here $\oplus$ denotes concatenation along the feature dimension.
\par\indent
Minimizing $\mathcal{L}_{rec}$ encourages $H_s^m$ and $H_c^m$ to jointly preserve the input information, which stabilizes disentanglement and reduces trivial solutions.

\noindent \textbf{(2) Independence via Orthogonality.}
To eliminate redundancy between the two subspaces, we enforce geometric orthogonality. By minimizing the Frobenius norm of the correlation between $H_s^m$ and $H_c^m$, we ensure that the two representations are linearly independent. This forces the model to isolate modality-specific noise within $H_s^m$, leaving $H_c^m$ to focus on shared semantics:
\begin{equation}
    \mathcal{L}_{orth} = \sum_{m \in \{t, a, v\}} \| (H_s^m)^\top H_c^m \|_F^2.
\end{equation}

\begin{figure}[t]
    \centering
    \includegraphics[width=\linewidth]{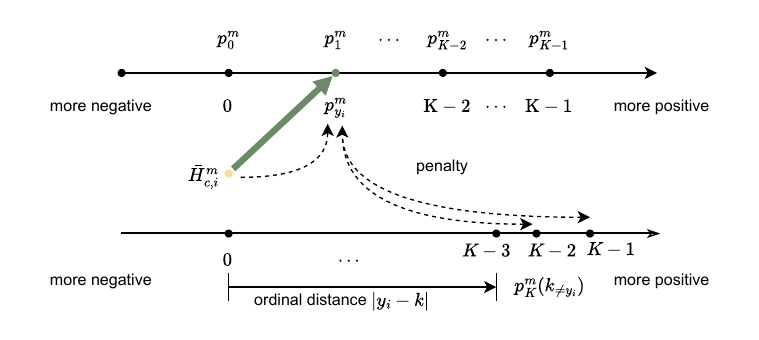} 
    \caption{\textbf{Illustration of the O-PCL.} 
    For modality $m$, the sample representation $\bar{H}_{c,i}^m$ serves as an anchor and is pulled toward its positive prototype $p_{y_i}^m$ (green arrow). 
    Crucially, negative prototypes $p_k^m$ ($k \neq y_i$) are pushed away with a dynamic penalty proportional to the ordinal distance $w_{i,k} = |y_i - k|/(K-1)$. 
    This is implemented by increasing the logits with $\alpha w_{i,k}$, forcing the latent space to respect the hierarchy of sentiment intensities.}
    \label{fig:opcl}
\end{figure}

\noindent \textbf{(3) Alignment via Ordinal Prototype Optimization.}
Sentiment intensity naturally follows an ordinal structure, where adjacent levels are semantically closer than distant ones.
For instance, confusing an extremely positive sample with an extremely negative one should incur a larger penalty than confusing it with a nearby positive level.
However, conventional prototype-based contrastive learning typically treats all negative classes uniformly, thereby overlooking this ordinal prior.
To address this, we propose an O-PCL, as illustrated in Figure \ref{fig:opcl}, which injects an explicit distance-dependent penalty into the prototype contrast objective.\par\indent

We obtain the ordinal label $y_i$ by quantizing the continuous sentiment score $\tilde{y}_i$ into $K$ ordered levels, which is consistent with the standard Acc-5 and Acc-7 evaluation protocol.
We denote the resulting ordinal label of sample $i$ as $y_i$,
\begin{equation}
y_i \in \{0,1,\dots,K-1\},
\end{equation}
where a larger $y_i$ indicates a more positive sentiment.
Accordingly, we define $Y_{\min}=0$ and $Y_{\max}=K-1$, so the maximum ordinal gap is $Y_{\max}-Y_{\min}=K-1$.\par\indent

Consistent with our implementation, we compute in-batch prototypes on-the-fly instead of introducing additional learnable prototype parameters.
Let $H_{c,i}^m$ be the common representation of sample $i$ in modality $m\in\{t,a,v\}$.
We first normalize each feature and then compute a prototype for each ordinal level that appears in the current mini-batch,
\begin{align}
\bar{H}_{c,i}^m &= \frac{H_{c,i}^m}{\|H_{c,i}^m\|_2}, \\
\mathcal{B}_k &= \{ i \mid y_i = k \}, \\
p_k^m &= \frac{1}{|\mathcal{B}_k|}\sum_{i\in\mathcal{B}_k}\bar{H}_{c,i}^m, \\
p_k^m &\leftarrow \frac{p_k^m}{\|p_k^m\|_2},
\end{align}
where $\mathcal{B}_k$ denotes the set of samples in the batch with ordinal label $k$.
In practice, we only construct prototypes for labels present in the batch, i.e., $k \in \mathcal{K}_{\mathcal{B}}=\{k \mid |\mathcal{B}_k|>0\}$.\par\indent

For each sample $i$ and prototype index $k$, we define the normalized ordinal distance factor,
\begin{equation}
w_{i,k}=\frac{|y_i-k|}{Y_{\max}-Y_{\min}}=\frac{|y_i-k|}{K-1},
\end{equation}
so that $w_{i,k}\in[0,1]$. This normalization keeps the penalty scale comparable across different choices of $K$.\par\indent

\textbf{Penalty-injected prototype contrast.}
We compute similarity logits between each sample representation and the in-batch prototypes, and inject the ordinal penalty only for negative prototypes,
\begin{align}
z_{i,k}^m &= \frac{\mathrm{sim}\!\left(\bar{H}_{c,i}^m,\, p_k^m\right)}{\tau}, \\
\tilde{z}_{i,k}^m &= z_{i,k}^m + \alpha\ \cdot  w_{i,k}\cdot \mathbb{I}[k\neq y_i],
\end{align}
where $\mathrm{sim}(\cdot,\cdot)$ is cosine similarity, $\tau$ is the temperature, and $\alpha$ controls the penalty strength.
Finally, we optimize a cross-entropy objective over prototypes,
\begin{align}
\mathcal{L}_{proto}^m
&= -\frac{1}{B}\sum_{i=1}^{B}
\log \frac{\exp(\tilde{z}_{i,y_i}^m)}
{\sum_{k\in\mathcal{K}_{\mathcal{B}}} \exp(\tilde{z}_{i,k}^m)}, \\
\mathcal{L}_{proto} &= \sum_{m\in\{t,a,v\}} \mathcal{L}_{proto}^m.
\end{align}
\par\indent
By enlarging the logits of ordinal-distant negative prototypes through $\alpha w_{i,k}$, O-PCL increases the loss for confusing emotionally distant levels, thereby pushing representations away from distant prototypes and shaping an ordinal-structured latent space.\par\indent

\subsection{Text-Guided Hybrid MoE}
\label{sec:maestro}
\begin{figure}[t]
  \centering
  \includegraphics[width=1.0\linewidth]{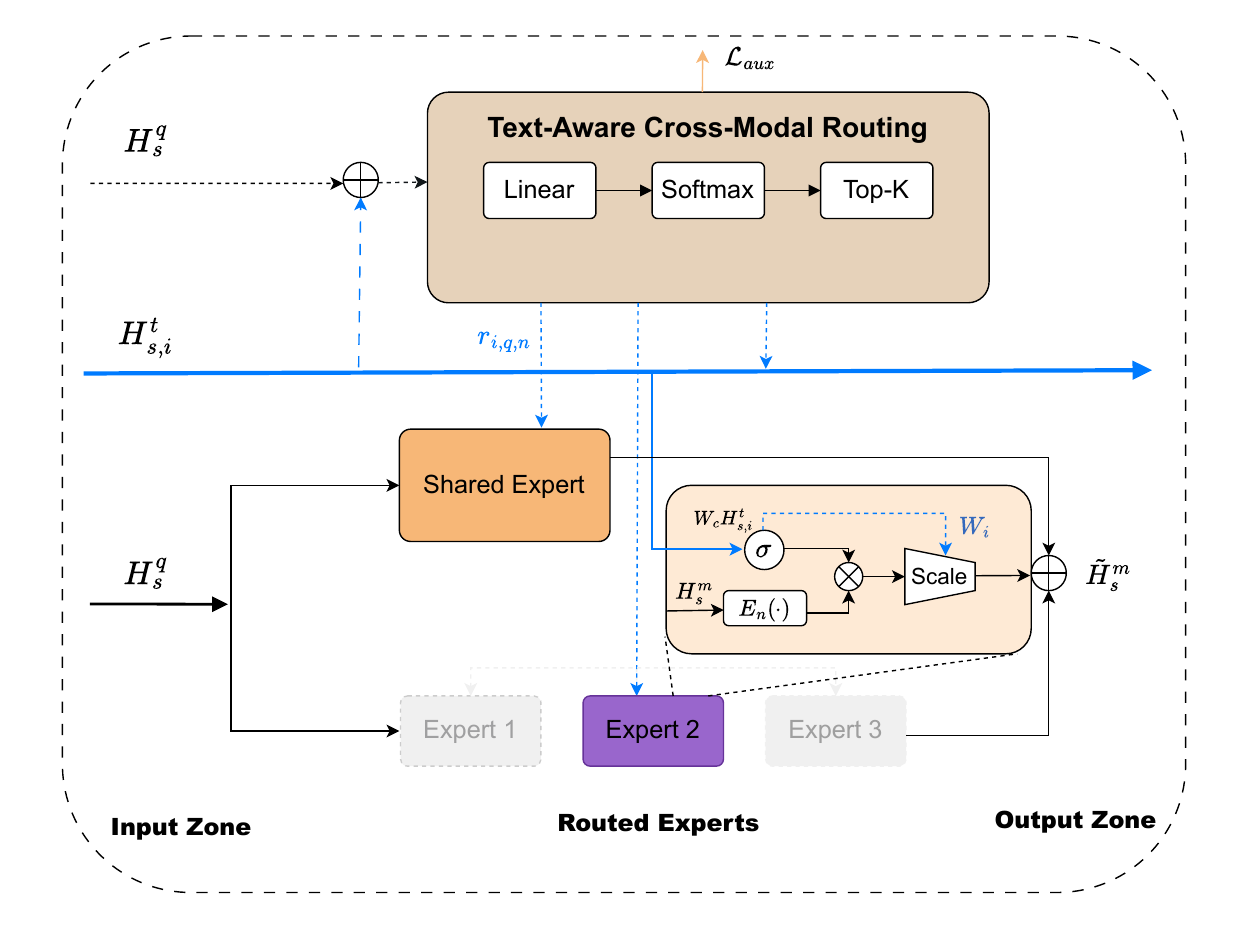}
  \caption{Illustration of the Maestro Block, featuring Text-Guided Dynamic Routing and Context-Aware Dual Gating to orchestrate non-verbal representations.}
  \label{fig:maestro_detail}
\end{figure}

The interpretation of non-verbal cues is highly context dependent. A visual smile may indicate happiness or sarcasm, and the correct meaning often depends on the linguistic content. Standard MoE methods typically adopt a self-routing strategy, where expert selection is conditioned only on the input modality itself. As a result, the modality is processed in isolation from the semantic context, which resembles an orchestra performing without a conductor. We address this limitation by introducing Maestro, as illustrated in Figure \ref{fig:maestro_detail}. Maestro uses the text-specific feature $H_s^t$ as a global conductor signal and dynamically orchestrates the processing of the audio feature $H_s^a$ and the visual feature $H_s^v$.

\subsubsection{Text-Guided Dynamic Routing}
\par
The conductor selects a sparse set of experts to process the current sample in a context-aware manner. Unlike standard MoE designs that perform self-routing within each modality, we introduce a text-guided cross-modal router. The key idea is to condition expert selection for non-text modalities on both the global linguistic context and the local modality dynamics.

For a target non-text modality $q \in \{a, v\}$, we compute routing logits over $N$ experts for each sample $i$,
\begin{equation}
s_{i,q} = W_g \left( H_{s,i}^t \oplus H_{s,i}^q \right) + b_g, \qquad s_{i,q} \in \mathbb{R}^{N},
\end{equation}
where $H_{s,i}^t \in \mathbb{R}^{d_t}$ and $H_{s,i}^q \in \mathbb{R}^{d_q}$ are the text-specific and modality-specific representations of sample $i$ after temporal aggregation, $W_g \in \mathbb{R}^{N \times (d_t + d_q)}$ and $b_g \in \mathbb{R}^{N}$ are learnable parameters.
\par\indent
The gating distribution is obtained by
\begin{equation}
G_{i,q} = \mathrm{Softmax}(s_{i,q}), \qquad G_{i,q} \in \mathbb{R}^{N}.
\end{equation}
\par\indent
We adopt Top-$K$ routing and activate only the $K$ experts with the largest gating scores. The sparse routing weight assigned to expert $n$ is
\begin{equation}
r_{i,q,n} =
\begin{cases}
\frac{G_{i,q,n}}{\sum_{j \in \mathrm{TopK}(G_{i,q},K)} G_{i,q,j}}, & n \in \mathrm{TopK}(G_{i,q},K), \\
0, & \text{otherwise}.
\end{cases}
\end{equation}
\par\indent
This sparse selection encourages expert specialization and filters irrelevant signals. We further apply an auxiliary load balancing objective during training to mitigate expert collapse, and we detail this objective in Section~\ref{sec:optimization}.

\subsubsection{Context-Aware Dual Gating}
Beyond expert selection, fine-grained modulation inside each expert is also crucial. We introduce a context-aware dual gating mechanism. The sparse routing weight $r_{i,q,n}$ controls the contribution of expert $n$, while an internal channel gate modulates the expert output using the text context.
For sample $i$, modality $q \in \{a, v\}$, and expert $n$, we compute
\begin{equation}
O_{i,q,n} = r_{i,q,n}\, \Big( E_n(H_{s,i}^q) \odot \sigma(W_c H_{s,i}^t) \Big),
\end{equation}
Here, $E_n$ denotes the $n$-th expert network, $\sigma$ is the Sigmoid activation function, and $\odot$ denotes element-wise multiplication. $W_c \in \mathbb{R}^{d_t \times d_E}$ projects the text-specific feature $H_{s,i}^T \in \mathbb{R}^{d_t}$ to the expert channel space of dimension $d_E$.
\par\indent

\subsubsection{Hybrid Expert Aggregation}
\par
Purely sparse MoE models may suffer from training instability. To balance dynamic adaptability with static robustness, we adopt a hybrid aggregation scheme. For sample $i$ and modality $q \in \{a, v\}$, the enhanced representation is
\begin{equation}
\tilde{H}_{s,i}^q = E_{shared}^q(H_{s,i}^q) + \sum_{n=1}^{N} O_{i,q,n}.
\end{equation}
\par\indent
Since $r_{i,q,n}=0$ for non-selected experts, the summation non-zero only for Top-K routed experts over the Top-$K$ routed experts.
\par\indent
Here, the shared expert captures modality-invariant patterns that are useful across contexts, while the routed experts provide text-aligned refinements.
\par\indent

\subsection{Multimodal Fusion and Prediction}
After the text-guided enhancement via the Maestro Block, we obtain the refined modality-specific features $\tilde{H}_s^a$ and $\tilde{H}_s^v$, which now incorporate linguistically aligned non-verbal cues. For simplicity, we omit the sample index $i$ in the following fusion and prediction formulas.
Simultaneously, we retain the original text-specific feature $H_s^t$ and the modality-invariant common features $\{H_c^t, H_c^a, H_c^v\}$ extracted from the disentanglement module.

To condense the shared semantic information, we perform an average pooling operation over the common subspaces. This operation aggregates the modality-invariant features across the three modalities, ensuring that the most representative information from all modalities is captured in $H_c^{avg}$:
\begin{equation}
    H_c^{avg} = \frac{1}{3} \left( H_c^t + H_c^a + H_c^v \right).
\end{equation}
\par\indent
Subsequently, we construct a comprehensive multimodal representation vector $Z$ by concatenating the consolidated common feature with the specificity-enhanced representations:
\begin{equation}
    Z = H_c^{avg} \oplus H_s^t \oplus \tilde{H}_s^a \oplus \tilde{H}_s^v,
\end{equation}
where $Z \in \mathbb{R}^{d_{total}}$ encapsulates both the consistent sentiment semantics and the context-sensitive unimodal dynamics.

Finally, the fused vector $Z$ is fed into a sentiment regression head. This head consists of a multi-layer perceptron with two fully connected layers, non-linear activation functions, and dropout regularization to prevent overfitting. The predicted sentiment intensity $\hat{y}$ is computed as:
\begin{equation}
    \hat{y} = W_{2} \left( \mathrm{ReLU} \left( W_{1} Z + b_{1} \right) \right) + b_{2},
\end{equation}
where $W_{1}, W_{2}$ and $b_{1}, b_{2}$ represent the learnable weights and biases of the projection layers, respectively.

\subsection{Optimization Objectives}
\label{sec:optimization}
The MAESTRO framework is optimized in an end to end manner. The total objective is a weighted combination of the task loss and multiple regularization terms:
\begin{equation}
\mathcal{L}_{total} = \mathcal{L}_{task} + \lambda_1 \mathcal{L}_{orth} + \lambda_2 \mathcal{L}_{rec} + \lambda_3 \mathcal{L}_{proto} + \lambda_4 \mathcal{L}_{aux},
\end{equation}
where $\lambda_1, \lambda_2, \lambda_3, \lambda_4$ balance the contribution of each component.
\par\indent

Sentiment prediction on benchmarks such as CMU-MOSI is formulated as a regression problem. We use mean absolute error as the task objective:
\begin{equation}
\mathcal{L}_{task} = \frac{1}{B}\sum_{i=1}^{B}\left| \tilde{y}_i - \hat{y}_i \right|,
\end{equation}
where $B$ is the batch size, $\tilde{y}_i$ is the ground truth sentiment score, and $\hat{y}_i$ is the predicted score.
\par\indent

A key challenge in training sparse MoE models is expert collapse, where the router consistently activates only a small subset of experts and leaves others underutilized. To mitigate this issue during training, we introduce an auxiliary load balancing loss that encourages balanced expert utilization.

For a batch of $B$ samples and $N$ experts, let $f_n$ denote the fraction of samples routed to expert $n$, and let $P_n$ denote the average gating probability assigned to expert $n$ over the batch. We define
\begin{equation}
\mathcal{L}_{aux} = N \sum_{n=1}^{N} f_n \cdot P_n,
\end{equation}
which is optimized together with the main objective to discourage skewed routing.
\par\indent
This auxiliary term complements Top-$K$ sparse routing by preventing a few experts from dominating the traffic, while still allowing the router to focus computation on the most informative experts for each sample.
\par\indent

\section{EXPERIMENTS}
\label{sec:experiments}

\subsection{Experiment Setup}

\noindent \textbf{Datasets.} To validate the effectiveness of our framework, we conduct experiments on two standard benchmarks, namely CMU-MOSI \cite{zadeh2016mosi} and its larger counterpart CMU-MOSEI \cite{zadeh2018mosei}. The former comprises 2,199 video segments, whereas the latter features a more extensive collection of 22,856 clips with enhanced speaker diversity. All samples across both datasets are annotated with continuous sentiment intensity scores where -3 represents strongly negative and +3 indicates strongly positive sentiments.

\textbf{Implementation Details.} All experiments were conducted on a single NVIDIA RTX 4090 GPU. We employed \texttt{bert-base-uncased} as the textual encoder. To prevent overfitting on limited multimodal data, we projected all modality features into a compact common subspace. For the acoustic and visual modalities, we utilized 2-layer Transformer backbones \cite{vaswani2017attention}. The model was optimized using the Adam optimizer with a unified learning rate of $1e^{-4}$. To stabilize the MoE training, we applied gradient clipping with a threshold of 0.6. The batch size was set to 64, and we employed early stopping with a patience of 5 epochs. Key hyperparameters including dataset-specific dropout rates are detailed in Table \ref{tab:hyperparams}.

\begin{table}[h]
\centering
\caption{Hyperparameter settings for CMU-MOSI and CMU-MOSEI.}
\label{tab:hyperparams}
\resizebox{0.95\linewidth}{!}{
\begin{tabular}{lcc}
\hline
\textbf{Parameter} & \textbf{CMU-MOSI} & \textbf{CMU-MOSEI} \\ \hline
\multicolumn{3}{l}{\textit{Training Config}} \\
Batch size & 64 & 64 \\
Learning Rate & $1e^{-4}$ & $1e^{-4}$ \\
Gradient Clip & 0.6 & 0.6 \\
Patience & 5 & 5 \\
\hline
\multicolumn{3}{l}{\textit{Model Architecture}} \\
Transformer Layers & 2 & 2 \\
Num. Experts ($N$) & 4 & 4 \\
Top-K ($K$) & 2 & 2 \\
\hline
\multicolumn{3}{l}{\textit{Regularization}} \\
Text Dropout & 0.5 & 0.1 \\
Attn Dropout & 0.3 & 0.5 \\
$\lambda_{orth}$ (Orthogonality) & 0.1 & 0.1 \\
$\lambda_{aux}$ (Load Balance) & 0.1 & 0.1 \\
$\alpha$ (O-PCL Penalty) & 0.5 & 0.5 \\ \hline  
\end{tabular}
}
\end{table}

\textbf{Metrics.} Following previous studies, we evaluate performance using six metrics: Mean Absolute Error (MAE), Pearson correlation coefficient (Corr), binary classification accuracy (Acc-2), F1 score (F1), 5-class accuracy (Acc-5), and 7-class accuracy (Acc-7). Higher Acc/F1/Corr and lower MAE indicate better performance.

\subsection{Results and Analysis}
\label{sec:main_results}

\noindent \textbf{Baselines.} To comprehensively evaluate the effectiveness of MAESTRO, we compare it against a wide range of state-of-the-art MSA methods categorized into two groups: (1) Fusion and Attention-based approaches, including TFN \cite{zadeh2017tensor}, LMF \cite{liu2018efficient}, MulT \cite{tsai2019multimodal}, and MAG-BERT \cite{rahman2020integrating}; and (2) Advanced Representation Learning frameworks, which cover disentanglement and contrastive learning methods such as MISA \cite{hazarika2020misa}, Self-MM \cite{yu2021learning}, HyCon \cite{mai2022hybrid}, ConFEDE \cite{yang2023confede}, DMD \cite{li2023decoupled}, and DLF \cite{wang2025dlf}.

\textbf{Performance Comparison.} The quantitative results on CMU-MOSI and CMU-MOSEI are summarized in Table \ref{tab:main_results}. MAESTRO establishes a new state-of-the-art by outperforming strong baselines across most key metrics.

\begin{table*}[t]
\centering
\caption{Comparison with state-of-the-art methods on CMU-MOSI and CMU-MOSEI. The best results are highlighted in \textbf{bold}. The suffix $^*$ indicates results cited from related literature.}
\label{tab:main_results}
\resizebox{\textwidth}{!}{
\begin{tabular}{lcccccccccccc}
\toprule
\multirow{2}{*}{\textbf{Method}} & \multicolumn{6}{c}{\textbf{CMU-MOSI}} & \multicolumn{6}{c}{\textbf{CMU-MOSEI}} \\
\cmidrule(lr){2-7} \cmidrule(lr){8-13}
 & \textbf{Acc-7} $\uparrow$ & \textbf{Acc-5} $\uparrow$ & \textbf{Acc-2} $\uparrow$ & \textbf{F1} $\uparrow$ & \textbf{Corr} $\uparrow$ & \textbf{MAE} $\downarrow$ & \textbf{Acc-7} $\uparrow$ & \textbf{Acc-5} $\uparrow$ & \textbf{Acc-2} $\uparrow$ & \textbf{F1} $\uparrow$ & \textbf{Corr} $\uparrow$ & \textbf{MAE} $\downarrow$ \\ 
\midrule
TFN & 34.90 & 39.39 & 80.08 & 80.07 & 0.698 & 0.901 & 50.20 & 53.10 & 82.50 & 82.10 & 0.700 & 0.593 \\
LMF & 33.20 & 38.13 & 82.50 & 82.40 & 0.695 & 0.917 & 48.00 & 52.90 & 82.00 & 82.10 & 0.677 & 0.623 \\
MulT & 40.00 & 42.68 & 83.00 & 82.00 & 0.698 & 0.871 & 51.80 & 54.18 & 82.50 & 82.30 & 0.703 & 0.580 \\
MISA & 41.37 & 47.08 & 83.54 & 83.58 & 0.778 & 0.777 & 52.05 & 53.63 & 84.67 & 84.66 & 0.752 & 0.558 \\
MAG-BERT & 43.62 & - & 84.43 & 84.61 & 0.781 & 0.727 & 52.67 & - & 84.82 & 84.71 & 0.755 & 0.543 \\
HyCon$^*$ & 46.60 & - & 85.20 & 85.10 & 0.790 & 0.713 & 52.80 & - & 85.40 & 85.60 & 0.776 & 0.601 \\
Self-MM$^*$ & 46.67 & - & 85.46 & 85.43 & 0.796 & 0.708 & 53.87 & - & 85.15 & 84.90 & 0.765 & 0.531 \\
ConFEDE$^*$ & 42.27 & - & 85.52 & 85.52 & 0.784 & 0.742 & 54.86 & - & 85.82 & 85.83 & 0.780 & 0.522 \\
DMD$^*$ & 46.06 & - & 83.23 & 83.29 & - & 0.752 & 52.78 & - & 84.62 & 84.62 & - & 0.543 \\
DLF & 47.08 & 52.33 & 85.06 & 85.04 & 0.781 & 0.731 & 53.90 & 55.70 & 85.42 & 85.27 & 0.764 & 0.536 \\ 
\midrule
\textbf{MAESTRO (Ours)} & \textbf{49.42} & \textbf{56.56} & \textbf{87.20} & \textbf{87.16} & \textbf{0.807} & \textbf{0.689} & \textbf{54.54} & \textbf{56.28} & \textbf{85.91} & \textbf{85.88} & \textbf{0.769} & \textbf{0.529} \\ 
\bottomrule
\end{tabular}
}
\end{table*}

\textbf{Analysis on CMU-MOSI.} As shown in Table \ref{tab:main_results}, MAESTRO demonstrates superior performance over the strongest competitor, ConFEDE. Specifically, our model attains an Acc-2 score of \textbf{87.20\%}, outperforming the strongest baseline ConFEDE by a clear margin. Most notably, we observe a substantial improvement in regression precision which validates the effectiveness of our Ordinal-Aware approach. Compared to ConFEDE and Self-MM, MAESTRO attains a lower MAE of \textbf{0.689} and achieves a superior correlation of \textbf{0.807}. While recent methods like ConFEDE effectively fuse modalities, they often neglect the ordinal distance between sentiment labels. By integrating the O-PCL, MAESTRO explicitly penalizes large sentiment deviations such as confusing +3 with -1, forcing the latent space to respect sentiment hierarchy. This structural alignment directly translates into the observed lower MAE and higher correlation.

\textbf{Analysis on CMU-MOSEI.} On the large-scale MOSEI dataset, MAESTRO continues to show competitive advantages. It surpasses all baselines in classification tasks, achieving the highest Acc-2 of \textbf{85.91\%} and F1 score of \textbf{85.88\%}, edging out the strong baseline ConFEDE. The consistent performance on MOSEI attributes to the Text-Guided Dynamic Routing. Unlike static fusion mechanisms used in DLF or Self-MM that apply fixed interactions, MAESTRO acts as a "conductor" by dynamically activating only the most semantically relevant audio-visual experts for each sample. This adaptability ensures robustness even in the diverse and noisy environments characteristic of large-scale datasets.

\subsection{Ablation Study}
\label{sec:ablation}

To thoroughly investigate the contribution of each core component within the MAESTRO framework, we conducted a comprehensive ablation study on the CMU-MOSI dataset. We constructed four distinct variants to isolate the effects of the dynamic MoE mechanism, the text-guided routing strategy, the feature disentanglement module, and the ordinal-aware optimization:

\begin{itemize}
    \item \textbf{w/o Maestro Block}: In this setting, we remove the entire Maestro MoE structure. Instead of dynamically selecting experts, the model utilizes a standard static fusion approach where features from text, audio, and video encoders are directly concatenated. This setting serves as a baseline to quantify the value of dynamic computation over static interaction.
    
    \item \textbf{w/o Text-Guided Router}: This variant retains the MoE architecture but replaces the Text-Guided Router with a Self-Routing mechanism. Specifically, the audio and visual experts are selected based solely on their own modality features, without the "conductor" signal from the textual context. This is designed to test whether linguistic guidance is superior to self-adaptation in resolving cross-modal ambiguity.
    \item \textbf{w/o Disentanglement}: We remove the modality-specific and modality-invariant decomposition along with the orthogonality and reconstruction constraints. The raw encoded features are directly fed into the Maestro Block. This setting tests whether purifying modality dynamics is a prerequisite for effective expert routing.
    \item \textbf{w/o O-PCL}: We replace the proposed O-PCL with a standard contrastive loss that treats all negative samples equally. The model is trained without the distance-based penalty weights. This setup evaluates whether incorporating ordinal priors helps the model focus on fine-grained sentiment intensity and prevents regression errors.
\end{itemize}

The results presented in Table \ref{tab:ablation} offer several critical insights into the architecture of MAESTRO.

\begin{table}[h]
\centering
\caption{Ablation study results of MAESTRO on CMU-MOSI. We report six metrics to fully evaluate the performance impact. \textbf{Bold} indicates the best performance.}
\label{tab:ablation}
\resizebox{\linewidth}{!}{
\begin{tabular}{lcccccc}
\toprule
\textbf{Model} & \textbf{Acc-7} & \textbf{Acc-5} & \textbf{Acc-2} & \textbf{F1} & \textbf{MAE} & \textbf{Corr} \\ 
\midrule
w/o Maestro Block & 45.32 & 52.14 & 84.75 & 84.71 & 0.738 & 0.789 \\
w/o Text-Guided Router & 46.85 & 54.02 & 85.42 & 85.38 & 0.719 & 0.795 \\
w/o Disentanglement & 46.10 & 53.50 & 85.15 & 85.12 & 0.728 & 0.792 \\
w/o O-PCL & 47.90 & 55.10 & 85.80 & 85.76 & 0.724 & 0.798 \\ 
\midrule
\textbf{MAESTRO (Full)} & \textbf{49.42} & \textbf{56.56} & \textbf{87.20} & \textbf{87.16} & \textbf{0.689} & \textbf{0.807} \\ 
\bottomrule
\end{tabular}
}
\end{table}

\textbf{Dynamic Selection Outperforms Static Fusion.} The "w/o Maestro Block" variant exhibits the lowest performance across all metrics with the Acc-7 dropping significantly to 45.32\%. This confirms that static computation graphs are insufficient for capturing complex multimodal dynamics. Simply concatenating features introduces redundancy and noise. In contrast, the Maestro Block effectively filters out irrelevant information through dynamic routing, allowing the model to focus on the most informative cues for each specific sample.

\textbf{Linguistic Context Disambiguates Non-Verbal Cues.} The "w/o Text-Guided Router" improves upon the static baseline but lags behind the full MAESTRO, recording an Acc-2 of 85.42\% compared to 87.20\%. This supports our hypothesis that non-verbal signals are often ambiguous without semantic context such as when a smile indicates sarcasm rather than happiness. The Text-Guided Router acts as a "conductor" ensuring that the activated visual and acoustic experts are semantically aligned with the linguistic intent, thereby guaranteeing interpretation reliability.

\textbf{Disentanglement Ensures Routing Reliability.} The performance drop observed in the "w/o Disentanglement" variant (Acc-2 drops to 85.15\%) highlights the foundational role of feature purification. Without disentanglement, modality-specific features are polluted by irrelevant noise and shared semantics. This entanglement confuses the Text-Guided Router, making it difficult to accurately assess which expert is needed. This result confirms that "clean inputs make for a better conductor," validating the necessity of our geometry-guided disentanglement module.

\textbf{Ordinal Constraints Enhance Regression Precision.} Notably, the "w/o O-PCL" variant suffers a clear degradation in regression performance with MAE increasing from 0.689 to 0.724. This is a crucial finding because standard contrastive losses ignore the semantic distance between labels and fail to penalize ordinal errors effectively. The O-PCL ensures that the latent space is structured according to sentiment intensity, enabling the model to distinguish fine-grained emotional shifts such as +3 versus +2 and achieve high-precision predictions.

\subsection{Further Analysis}
\label{sec:further_analysis}

To provide deeper insights into the interpretability of MAESTRO, we investigate how the Text-Guided Router orchestrates multimodal interactions by visualizing the dynamic routing weights. 

We conducted a case study on a representative sample from the CMU-MOSI test set involving sarcasm where the speaker says ``\textit{It's just... fine}'' with a disappointed facial expression. In this scenario, the literal positive meaning of the text contradicts the negative non-verbal signals, often confusing static fusion models.

\begin{figure}[htbp]
  \centering
  \includegraphics[width=0.95\linewidth]{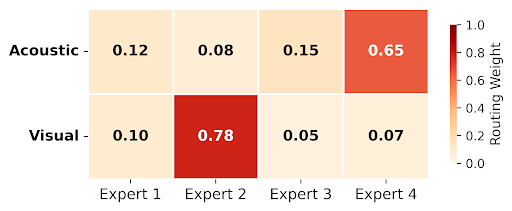} 
  \Description{A heatmap showing the routing weights of four experts across acoustic and visual modalities.}
  \caption{Visualization of dynamic routing weights. The heatmap illustrates that the Text-Guided Router resolves semantic ambiguity by assigning dominant weights to Visual Expert 2 and Acoustic Expert 4, as indicated by the dark red intensity.}
  \label{fig:case_study}
\end{figure}

As illustrated in Figure \ref{fig:case_study}, the heatmap exhibits a distinct sparse activation pattern. Specifically, the router allocates dominant attention to Visual Expert 2 and Acoustic Expert 4, assigning them weights of 0.78 and 0.65, respectively, while effectively suppressing irrelevant experts. This behavior indicates that, guided by the linguistic cues of hesitation, the model adaptively redirects its focus towards facial expressions and vocal tonality to capture the underlying negative sentiment. This empirical evidence validates the capability of our text-driven "conductor" mechanism to resolve cross-modal ambiguity.

\section{Conclusion}
\label{sec:conclusion}

In this paper, we presented MAESTRO, a novel framework that shifts the paradigm of Multimodal Sentiment Analysis from static computation graphs to dynamic, context-aware routing. Drawing inspiration from an orchestra conductor, our proposed Text-Guided Router utilizes linguistic semantics to dynamically orchestrate audio-visual experts. This mechanism effectively resolves cross-modal ambiguity by activating only the most discriminative non-verbal cues for each specific sample. Furthermore, to address the neglect of sentiment hierarchy in existing methods, we introduced the O-PCL. By incorporating distance-based penalties into prototype learning, O-PCL enforces a structured latent space that rigorously aligns with the intrinsic order of sentiment intensities.

Extensive experiments on the benchmark datasets CMU-MOSI and CMU-MOSEI demonstrate that MAESTRO establishes a new state-of-the-art performance. Qualitative analysis further confirms the interpretability of our dynamic mechanism, revealing that the model intelligently shifts its focus to specific non-verbal signals when facing semantic conflicts such as sarcasm.
%%
%% The acknowledgments section is defined using the "acks" environment
%% (and NOT an unnumbered section). This ensures the proper
%% identification of the section in the article metadata, and the
%% consistent spelling of the heading.

%%
%% The next two lines define the bibliography style to be used, and
%% the bibliography file.
\bibliographystyle{ACM-Reference-Format}
\bibliography{refs}

%%
%% If your work has an appendix, this is the place to put it.
\appendix

\end{document}